\documentclass[runningheads]{llncs}
\usepackage[utf8]{inputenc}

\usepackage[linesnumbered,ruled,vlined]{algorithm2e}
\usepackage{amsmath}
\usepackage{amssymb}
\usepackage{tabularx}

\usepackage{graphicx}

\begin{document}

\title{The Mathematical Game}

\author{Marc Pierre \and
Quentin Cohen-Solal \and
Tristan Cazenave
}

\institute{
LAMSADE, Université Paris Dauphine - PSL, CNRS, Paris, France
}

\maketitle

\begin{abstract}
Monte Carlo Tree Search can be used for automated theorem proving. Holophrasm is a neural theorem prover using MCTS combined with neural networks for the policy and the evaluation. In this paper we propose to improve the performance of the Holophrasm theorem prover using other game tree search algorithms.

\end{abstract}

\section{Introduction}

Monte Carlo Tree Search (MCTS) has been successfully applied to many games and problems \cite{BrownePWLCRTPSC2012}. It was used to build superhuman game playing programs such as AlphaGo \cite{Silver2016MasteringTG}, AlphaZero \cite{silver2018general} and Katago \cite{wu2019accelerating}. It has been recently used to discover new fast matrix multiplication algorithms \cite{fawzi2022discovering}. It is also used for automated theorem proving such as in the Holophrasm theorem prover \cite{whalen2016holophrasm}. In this paper, we propose to replace the MCTS used by Holophrasm by other game tree search algorithms.  We will start by briefly explaining how metamath works, as well as the Holophrasm interface. Then we will move on to an application of existing tree search algorithms modified for this context, such as Minimax, PUCT or Product Propagation. Finally, we propose a new algorithm, which is an association of existing ones, and apply it in the context of Holophrasm.

\section{Holophrasm and Metamath}

\subsection{Metamath}
Metamath is a formal mathematics language. Its main function is based on the principle of logical substitution. For example, let's imagine that we are at a certain step of a theorem's proof, and to move on to the next step we need to apply a proposition. To do this, we need to change the variables of the proposition we wish to apply, so that its hypotheses correspond to the current state of the proof. For more details, one can check the metamath's book (\cite{metamathbook}). Holophrasm transforms this structure of theorem's proof into an "AND/OR" tree. OR nodes represents the current state of the proof. They are considered proven if one of their child is proven or if they are one of the initial hypotheses of the theorem. Their children are AND nodes representing a proposition to be applied to prove the OR father node. The children of an AND node are the set of hypotheses to be proven for the proposition to be true. A hypothesis is modeled by an OR node and an AND node is considered proven if all its children are proven. In Holophrasm, the theorem is proved by working backward. The root being the conclusion of the proof modeled by an OR node, each of its children is an AND node which is a proposition with its substitution of variables. Concerning the theorems on which we will test our algorithms, the benchmarks are made up of a list of theorems and are provided by the Holophrasm interface. However, due to time constraints, we will only test on the first 200 theorems in the list.
\subsection{Classical Holophrasm}
Now that we have seen the structure of proof trees, we will explain the search used by Holophrasm \cite{whalen2016holophrasm} through this trees.
\subsubsection{Algorithm}The algorithm visits the root using $VisitNodeOR$ as long as this root has not been proven or the number of his visits has not reached a certain threshold.
The value of the root is, as for all OR nodes, a probability calculated by the payout neural network modeling the chance of proving the OR node with the hypotheses it contains. When visiting an OR node, the objective is then to  visit an AND node (proposition), which has the highest chance of being provable. Note that for an AND node, its probability is given by the prediction neural network and the substitution is given by the generative network. Furthermore, when an AND node is created, all its children are also created and visited once.
Initial values are given by two different networks, depending on the nature of the node. The payout network takes an OR node as an argument and outputs the probability that this node is provable with the assumptions it contains. The prediction network gives a softmax on the set of propositions applicable to an OR node. Another feature we would like to detail is the interface adding an AND node to an OR node. Holophrasm uses a heap in which all potential candidates are stored. A visit to an OR node may not add an AND node, if the candidate is not valid. An OR node will be considered disproven if all its children are disproof, or if it has been visited enough times and the heap is empty. An AND node is disproven if one of its childs is disproven. Disproven AND node are cut from the tree. 

\begin{algorithm}
  \SetAlgoLined
  \SetKwFunction{Fonction}{Holophrasm}
  \SetKwProg{Fn}{Function}{:}{}
  \Fn{\Fonction{node, maxpasses}}{
    passes = 0\;
    \While{(not node.proven) or passes$<$maxpasses}{
    VisitNodeOR(node)\;
    UpdateProven(node, "OR")\;
    UpdateValueOR(node)\;
    }
  }
  \caption{Research Function of Holophrasm}
\end{algorithm}

\begin{algorithm}
  \SetAlgoLined
  \SetKwFunction{Fonction}{VisitNodeAND}
  \SetKwProg{Fn}{Function}{:}{}
  \Fn{\Fonction{node}}{    
    \If{$len(node.children)>0$}{
    VisiteNodeOR(node.children[argmin([child.value/child.visit for child in node.children])])\;
  }
    UpdateProven(node, "AND")\;
    UpdateValueAND(node)\;}
  \caption{Visit of an "AND" node by Holophrasm}
\end{algorithm}

\begin{algorithm}
  \SetAlgoLined
  \SetKwFunction{Fonction}{UpdateValueAND}
  \SetKwProg{Fn}{Function}{:}{}
  \Fn{\Fonction{node}}{    
    \If{$len(node.children)>0$}{
    badchild = node.children[argmin([child.value/child.visit for child in node.children])]\;
    node.visit = badchild.visit\;
    node.value = badchild.value\;
  }}
  return 0\;
  \caption{Update of an "AND" node by Holophrasm}
\end{algorithm}

\begin{algorithm}
  \SetAlgoLined
  \SetKwFunction{Fonction}{VisitNodeOR}
  \SetKwProg{Fn}{Fonction}{:}{}
  \Fn{\Fonction{node}}{   
    \If{node.heap is empty}{
        Use the Holophrasm Interface to fill node.heap of all compatible proposition\;
    }
    \If{$node.visit/6 + 0.01 > len(self.children)+node.childlessvisit$}{
      Try to add an "AND" node from the heap, if it fail node.childlessvisit+=1\;
      return 0\;
    }
    \If{$len(node.children)>0$}{
    value = 0\;
    nextchild = None\;
    \For{child $in$ node.children}{
    \If{valuation-function(node.visit, child) $>$ value}{
    value = valuation-function(node.visit, child)\;
    nextchild = child\;
    }
    VisiteNodeAND(nextchild)\;
    }
  }
    UpdateProven(node, "OR")\;
    UpdateValueOR(node)\;}
  \caption{Visit of an "OR" node by Holophrasm}
\end{algorithm}

\begin{algorithm}
  \SetAlgoLined
  \SetKwFunction{Fonction}{UpdateValueAND}
  \SetKwProg{Fn}{Function}{:}{}
  \Fn{\Fonction{node}}{    
    \If{$len(node.children)>0$}{
    node.visit = sum(child.visit for child in node.children) + node.childlessvisit + 1\;
    node.value = node.networkpayout + sum(child.value for child in node.children)\;
  }}
  \caption{Update of an "OR" node by Holophrasm}
\end{algorithm}

\begin{algorithm}
  \SetAlgoLined
  \SetKwFunction{Fonction}{UpdateProven}
  \SetKwProg{Fn}{Function}{:}{}  
  \Fn{\Fonction{node, type}}{
    \If{type $==$ "OR"}{
    \If{len(node.children) $==0$ $and$ node.visit$>1$ $and$ node.heap $is$ empty}{ node.dead = True \;}
    \For{$child$ $in$ $node.children$}{
    \If{$child.proven$}{node.proven = True\;}}
    }
    \If{type $==$ "AND"}{
    node.proven = True\;
    \For{$child$ $in$ $node.children$}{
    \If{$not$ $child.proven$}{
    node.proven = False\;
    break\;}}
    \If{$any$ ([child.dead $for$ child in node.children])}{CutNodeFromTree(node)\;return 0\;}}
  }
  \label{UpdateProven}
  \caption{Update if a node is proven}
\end{algorithm}

\begin{algorithm}
  \SetAlgoLined
  \SetKwFunction{Fonction}{valuation-function}
  \SetKwProg{Fn}{Function}{:}{}
  \Fn{\Fonction{fathervisit, nodeAND}}{
    return $\frac{nodeAND.value}{nodeAND.visit+1} + 0.5 * \frac{nodeAND.probabilitynet}{1 + nodeAND.visit} + \sqrt{\frac{\log(fathervisit)}{1+nodeAND.visit}}$\;
  }
  \caption{Evaluation Function for guiding the exploration in Holophrasm}
\end{algorithm}

\subsubsection{Results} 
\begin{figure}
  \centering
  \includegraphics[width=0.5\textwidth]{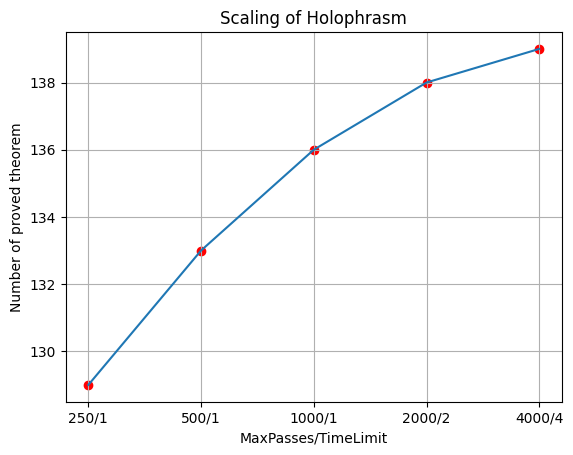}
  \caption{Results Holophrasm}
  \label{ScalingHolophrasm.png}
\end{figure}

On the first 200 theorems of the Holophrasm's Test set and with the parameter 10 for the BeamSearch used by the generative network, we obtain Figure \ref{ScalingHolophrasm.png}.\newline

We are not going to make a detailed analysis of the results, which we will use mainly to compare with the other algorithms we are going to test.

\section{ Classical Tree Search Algorithm for Metamath Theorem Proving}

In this section we will take a look at a number of well-known algorithms applied in this context by changing the Holophrasm search. We will start by analysing Minimax and its results, then we will study more selective algorithms such as PUCT.

\subsection{Minimax}

Now that we have tested the search algorithm provided with the Holophrasm interface, we are going to test a more traditional search: Minimax. The aim is to compare a more conventional search algorithm with the Holophrasm's results, and highlight the importance of progressive widening (the expansion of OR node's child).

\subsubsection{Algorithm}
We launch a Minimax search from the root by setting the depth. However, the problem is that an OR node can have an infinite number of children. To overcome this problem, we limit the number of possible children to a fixed breadth. The children of the OR node are chosen according to their probability given by the prediction neural network. The depth is reduced by 1 each time we go from an AND node to an OR node, and the algorithm ends on OR nodes. With regard to node initialization, when an AND node is created, all its OR child nodes are created and checked if they are an initial hypothesis of the problem. The point of testing several breadths is to evaluate the efficiency of the network in finding the next proposition to apply to an OR node.
\subsubsection{Results}
Our test set consists of the first 200 Holophrasm theorems. The test conditions are 1 pass in the maximum tree and a parameter of 10 for the BeamSearch used in the Holophrasm interface.\newline
\begin{tabularx}{\textwidth}{|>{\centering\arraybackslash}X |> {\centering\arraybackslash}X |> {\centering\arraybackslash}X |}
  \hline
   & depth = 2 & depth = 3  \\
  \hline
  breadth = 2 & 78/200 & 79/200   \\
  \hline
  breadth = 3  & 94/200 & 95/200\\
  \hline
  breadth = 4 & 97/200  & 98/200\\
  \hline
  breadth = 5 & 103/200 & 108/200\\
  \hline
  breadth = 7 & 112/200 & 113/200 \\
  \hline
  breadth = 9 & 117/200 & .../200 \\
  \hline
\end{tabularx}\newline\newline
From the results, we can see that the networks are often wrong, and that we need to go to wider breadths to get better results. This underlines the importance of the progressive widening used in Holophrasm.

\subsection{PUCT}

We now test other conventional algorithms such as PUCT \cite{Silver2016MasteringTG} , Product Propagation \cite{PPCazenaveSaffidine} and Proof Number Search \cite{ProofNumberSearch}. These tree search methods are more recent than Minimax and they allows one to find a solution without exploring the whole tree. Some of these approaches seams to not use networks, but in fact they are used implicitly by the interface when it attributes an And node to an Or node.

\subsubsection{Algorithm}
The idea behind PUCT is to supervise the search when choosing the next AND node to visit. This algorithm is inspired by the Bandit literature. To implement PUCT from Holophrasm's research, we will just change the Holophrasm bandit by changing the valuation-function.\newline
\begin{algorithm}[H]
  \SetAlgoLined
  \KwResult{PUCT}  
  \SetKwFunction{Fonction}{valuation-function}
  \SetKwProg{Fn}{Function}{:}{}
  \Fn{\Fonction{fathervisit, nodeAND}}{
    Return $\frac{nodeAND.value}{nodeAND.visit} + C * \frac{nodeAND.prediction\_net}{1 + nodeAND.visit} * \sqrt{fathervisit}$\;
  }
  \caption{Evaluation Function for PUCT}
\end{algorithm}

\subsubsection{Results}
\begin{figure}
  \centering
  \includegraphics[width=0.5\textwidth]{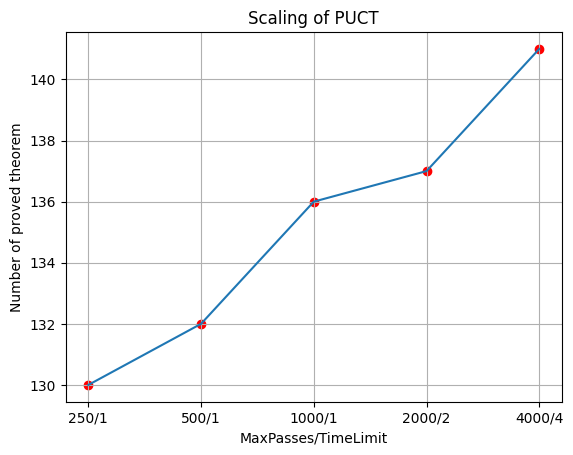}
  \caption{Results PUCT for C=0.2}
  \label{ScalingPUCT.png}
\end{figure}

First, we will determine the best constant to use in PUCT.
With the same BeamSearch setting as before and on the parameters (1000 passes, 1 min), we obtain:\newline
\begin{tabularx}{\textwidth}{|>{\centering\arraybackslash}X|>{\centering\arraybackslash}X|>{\centering\arraybackslash}X|>{\centering\arraybackslash}X|>{\centering\arraybackslash}X|>{\centering\arraybackslash}X|}
  \hline
   C  &0.1& 0.2 & 0.3  & 0.4 & 0.5  \\
  \hline
  Results&136/200& 136/200 & 136/200 & 136/200 & 135/200  \\
  \hline
\end{tabularx}
\newline\newline
We will therefore test the Scaling of PUCT with $C=0.2$, the results are described in Figure \ref{ScalingPUCT.png}. 

In comparison with Holophrasm, the results are quite similar except for the parameters (4000,4). So, with less time constraint, PUCT does better than Holophrasm. It is difficult to explain this behavior further. We will see below that there is a kind of ceiling around 140, whatever the algorithm used.
\subsection{Product Propagation}
Since PUCT is an algorithm that affects exploration, we will now look at Product Propagation (PP) \cite{PPCazenaveSaffidine}, which mainly changes the value and update of nodes.
\subsubsection{Algorithm}
The idea is to change the visit and the values attributed to the nodes during the search. The value is seen as a probability and the children are assumed to be independent. Thus the value of an AND node is the product of the values of its children and the value of an OR node is calculated using the same principle but with the additional probability. In an OR node we explore the best valued child, and in an AND node the worst. The value of Leaf can be initialized with 1 or by the payout network given by Holophrasm, but in our case we used the payout network for better results.\newline

\subsubsection{Results}
\begin{figure}
  \centering
  \includegraphics[width=0.5\textwidth]{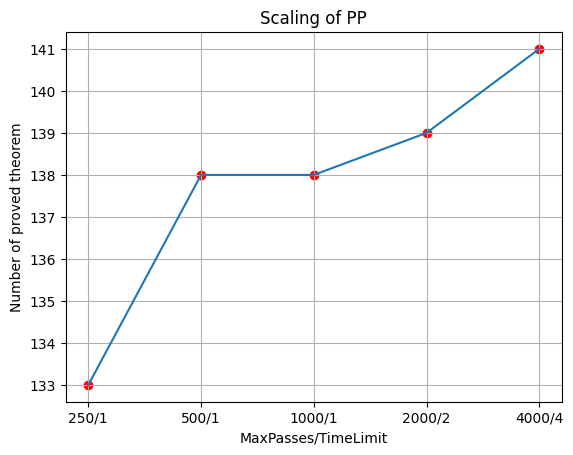}
  \caption{Results Product Propagation}
  \label{ScalingPP.png}
\end{figure}

With the the same settings as before we obtain Figure \ref{ScalingPP.png}.
The results show, in this context, that PP is more efficient than PUCT on parameters where the number of passes is limiting. However, for the parameter (4000,4), we achieve the same result as PUCT. PP seems to be more efficient, proving 138 propositions in just 1 minute and 500 passages maximum.

\subsection{Proof Number Search}
Similar to PP, Proof Number Search (PNS) \cite{ProofNumberSearch} is based on node values and updates. However, this time there are two values per node to take into account.
\subsubsection{Algorithm}
The idea is to count the number of leaves left to explore either to prove the node or to disprove it. During the node initialization, a non proven node is initialized with a Proof Number and a Disproof Number. In the original algorithm, for a non-proven or proven node, the PN and DPN are initialized to 1. For a proven node, the PN is set to 0 and the DPN to infinity, and vice versa for a proven node. Then during the visit, in an OR node we choose the AND node with the lowest DPN and in an AND node we choose the OR node with the lowest PN.  One might think that neural networks are not used in this algorithm, but they are used indirectly in AND node assignment.

\subsubsection{Results}
\begin{figure}
  \centering
  \includegraphics[width=0.5\textwidth]{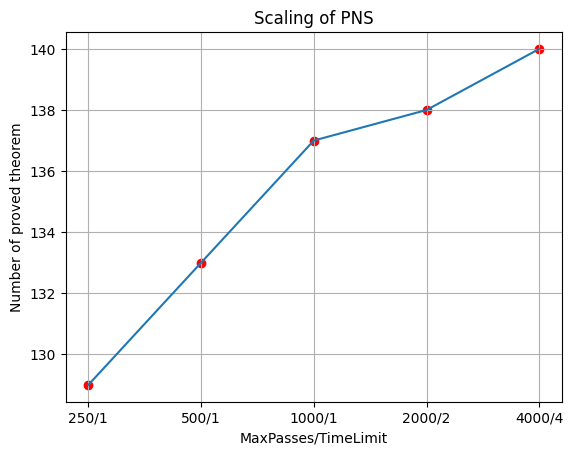}
  \caption{Results PNS}
  \label{ScalingPNS}
\end{figure}

With the same settings as before, we obtain Figure \ref{ScalingPNS}. The results are weaker than Product Propagation and all its variants (which we will see later), but in this test we are not using the value given by the "payout" neural network.
We tried to improve PNS, but we could not find any approach that improve the results presented above, either by initializing PN and DPN with the networks, or by using a PUCT in the search, so we haven't included these approaches in this article.

\subsection{HyperTree Proof Search}
We are now going to test an algorithm which is different in his approach from the previous ones. This algorithm try to obtain a broader perspective by selecting wider sub-tree instead of doing descent.
\subsubsection{Algorithm}
The goal is to draw inspiration from the search algorithm presented in \cite{HTP}, in order to compare the results of different approaches in this context. However, an adaptation is necessary because the interfaces used are different.  The idea is to select a sub-tree of the proof tree, expand the leaves, then update only the values of the AND nodes of this sub-tree using Product Propagation. Transposed to our interface, this is like using PUCT to select AND nodes, but once in an AND node, expand all its OR children.

\subsubsection{Results}
\begin{figure}
  \centering
  \includegraphics[width=0.5\textwidth]{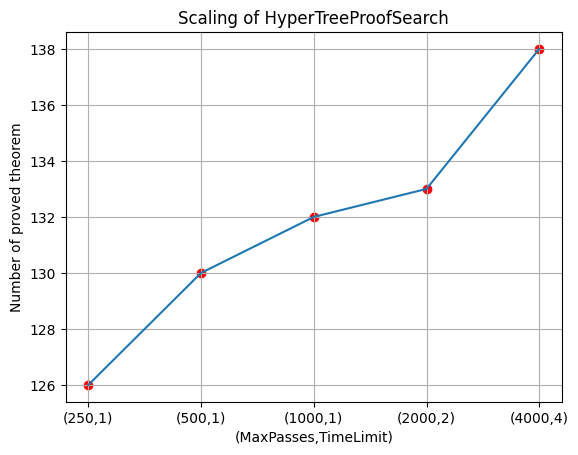}
  \caption{Results HyperTree Proof Search}
  \label{HTP.png}
\end{figure}

With 10 for the interface BeamSearch, we obtain the results described in Figure \ref{HTP.png}.

The results, in this context, are inferior to those of the previous algorithm. Extending all OR nodes that are children of an AND node is time-consuming. This explains the jump in performance from 2 to 4 minutes. This may be the algorithm with the worst results, but when you look at the evolution of the curve, it is the one that improves the most as a function of time.

\section{New Tree Search Algorithm for Metamath Theorem Proving}

Now that we have seen the more traditional approaches, we are going to use them as an inspiration to create new ones. 

\subsection{Production Propagation Combined with PUCT}

\subsubsection{Algorithm}
Given the performance of Product Propagation and PUCT, the idea is to combine the two approaches. In fact, these two algorithms should work well together, since Product Propagation does not rely on a bandit's part. We'll present the two combinations we've tested. For the first algorithm, the Product Propagation value is used, along with the PUCT visit and bandit. In the case of the second approach, we change the PUCT bandit (see algorithm \ref{modifiedBandit}). The aim was to keep the first Holophrasm bandit approach, but combine it with PUCT. This is useful especially when the policy given by the payout network is greatly underestimating a son. In this case the part coming from UCB compensates.\newline
\begin{algorithm}[H]
  \SetAlgoLined
  \SetKwFunction{Fonction}{modified-valuation-function}  
  \label{modifiedBandit}
  \SetKwProg{Fn}{Function}{:}{} 
  \Fn{\Fonction{fathervisit, nodeAND}}{
    Return $nodeAND.value $\newline$ + a * node.probabilitynet * \frac{\sqrt{fathervisit}}{nodeAND.visit}$\newline$ + b * \sqrt{\frac{\log(fathervisit)}{nodeAND.visit}}  $\;
  }
  \caption{Modified Evaluation function for PUCT}
\end{algorithm}
\subsubsection{Results}
With 10 for the interface BeamSearch and the modified validation function we obtain : \newline
\begin{tabularx}{\textwidth}{|>{\centering\arraybackslash}X|>{\centering\arraybackslash}X|>{\centering\arraybackslash}X|>{\centering\arraybackslash}X|>{\centering\arraybackslash}X|>{\centering\arraybackslash}X|>{\centering\arraybackslash}X|}
  \hline
   & a = 0.1  & a = 0.4 & a = 0.6 & a = 0.8 & a = 1.0 & a = 1.1\\
  \hline
  b = 0.1 & ... & 134/200 & ... & ... &... & ...\\
  \hline
  b = 0.4 & 133/200 & 135/200 & 136/200 && 137/200 & ... \\
  \hline
  b = 0.5 & ... & ... & ... & 138/200 & 137/200 & 136/200 \\
  \hline
  b = 0.6 & ... & 130/200 & 136/200 && 136/200 & ... \\
  \hline
  b = 1.0 & ... & ... & 134/200 && 135/200 & ... \\
  \hline
\end{tabularx}

\begin{figure}
  \begin{minipage}[t]{0.45\textwidth}
    \centering
    \includegraphics[width=\textwidth]{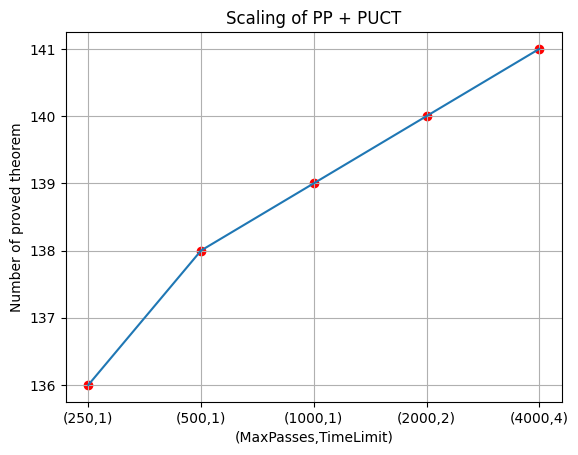}
    \caption{Results PP + PUCT c=0.2}
    \label{PPUCT}
  \end{minipage}\hfill
  \begin{minipage}[t]{0.45\textwidth}
    \centering
    \includegraphics[width=\textwidth]{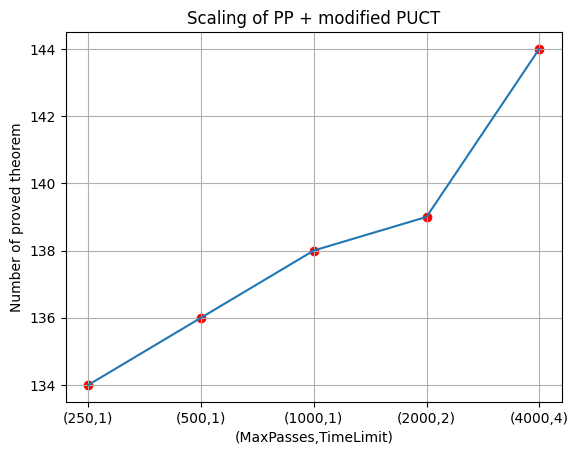}
    \caption{Results PP + PUCT with modified validation function a=0.8 b=0.5}
    \label{fig:image_droite}
  \end{minipage}
\end{figure}
In the case of the PUCT bandit (Figure \ref{PPUCT}), we can see that combining the two approaches results in a slight increase in performance. However, we are still unable to break the 141-theorem barrier.\par
There are two interesting points to note about the modified PUCT bandit (Figure \ref{fig:image_droite}). Firstly, the algorithm performs particularly well with parameters (4000,4), where it proves 144 theorems (and thus breaks the 141-theorem barrier). It is therefore the best on these criteria. Note that, this bandit exploration term has been used with other search algorithms, including Holophrasm, but its results were not convincing. The final point is that if we remove the principle of  node unprovability (see the UpdateProven \ref{UpdateProven} code line 3-4 and 20-22 ), this algorithm is the one with the best results.

\section{Conclusion}
We studied different algorithms applied to theorem proving in Metamath. Minimax algorithm highlighted the importance of progressive widening. Algorithms such as PUCT, PNS or PP obtained better results than the search used by Holophrasm. Finally, we propose a new algorithm by combining the idea of PUCT and PP and obtain better results.\par
Our initial goal was to improve the search used by the Holophrasm interface. The different algorithms we proposed, as well as the different solutions we provided, enabled this improvement. Although the improvement is slight, our different approaches seem more promising in terms of parameter-dependent changes in performance. In future work we plan to test this hypothesis by applying the algorithms on the whole dataset.

\bibliographystyle{splncs04}

\bibliography{main}

\begin{thebibliography}{10}
\providecommand{\url}[1]{\texttt{#1}}
\providecommand{\urlprefix}{URL }
\providecommand{\doi}[1]{https://doi.org/#1}

\bibitem{ProofNumberSearch}
Allis, L.V., van~der Meulen, M., van~den Henrik, H.: Proof-number search
  \textbf{1} (2003)

\bibitem{BrownePWLCRTPSC2012}
Browne, C., Powley, E., Whitehouse, D., Lucas, S., Cowling, P., Rohlfshagen,
  P., Tavener, S., Perez, D., Samothrakis, S., Colton, S.: A survey of {M}onte
  {C}arlo tree search methods. {IEEE} Transactions on Computational
  Intelligence and {AI} in Games  \textbf{4}(1),  1--43 (Mar 2012).
  \doi{10.1109/TCIAIG.2012.2186810}

\bibitem{fawzi2022discovering}
Fawzi, A., Balog, M., Huang, A., Hubert, T., Romera-Paredes, B., Barekatain,
  M., Novikov, A., R~Ruiz, F.J., Schrittwieser, J., Swirszcz, G., et~al.:
  Discovering faster matrix multiplication algorithms with reinforcement
  learning. Nature  \textbf{610}(7930),  47--53 (2022)

\bibitem{HTP}
Lample, G., Lachaux, M.A., Lavril, T., Martinet, X., Hayat, A., Ebner, G.,
  Rodriguez, A., Lacroix, T.: Hypertree proof search for neural theorem
  proving. Neural Information Processing Systems (NeurIPS)  (2022)

\bibitem{metamathbook}
Megill, N.D.: Metamath: A Computer Language for Mathematical Proofs. Lulu
  Press, Morrisville, North Carolina (2019)

\bibitem{PPCazenaveSaffidine}
Saffidine, A., Cazenave, T.: Developments on product propagation. In: CG 2013.
  pp. 100--109. Springer (2014)

\bibitem{Silver2016MasteringTG}
Silver, D., Huang, A., Maddison, C.J., Guez, A., Sifre, L., van~den Driessche,
  G., Schrittwieser, J., Antonoglou, I., Panneershelvam, V., Lanctot, M.,
  Dieleman, S., Grewe, D., Nham, J., Kalchbrenner, N., Sutskever, I.,
  Lillicrap, T., Leach, M., Kavukcuoglu, K., Graepel, T., Hassabis, D.:
  Mastering the game of go with deep neural networks and tree search. Nature
  \textbf{529},  484--489 (2016)

\bibitem{silver2018general}
Silver, D., Hubert, T., Schrittwieser, J., Antonoglou, I., Lai, M., Guez, A.,
  Lanctot, M., Sifre, L., Kumaran, D., Graepel, T., et~al.: A general
  reinforcement learning algorithm that masters chess, shogi, and go through
  self-play. Science  \textbf{362}(6419),  1140--1144 (2018)

\bibitem{whalen2016holophrasm}
Whalen, D.: Holophrasm: a neural automated theorem prover for higher-order
  logic. arXiv preprint arXiv:1608.02644  (2016)

\bibitem{wu2019accelerating}
Wu, D.J.: Accelerating self-play learning in go. arXiv preprint
  arXiv:1902.10565  (2019)

\end{thebibliography}

\end{document}